%% file: main.tex
\documentclass[10pt,conference,letterpaper]{IEEEtran}

\usepackage{algorithm}
\usepackage{algpseudocode}
\usepackage{amsmath}
\usepackage{amssymb}
\usepackage{amsthm}
\usepackage{enumitem}
\usepackage{xspace}
\usepackage{url}
\usepackage{booktabs}
\usepackage{graphicx}

\newtheorem{definition}{Definition}

\newcommand{\sysname}{\textsc{TWP}}
\newcommand{\itlmin}{\ensuremath{\ell_{\min}}}

\newcommand{\etal}{\textit{et al.}\xspace}

\begin{document}

\title{Stable Token-level Scheduling for Serverless LLM Inference}

\author{
\IEEEauthorblockN{Hardi Ashishbhai Desai, Balaji N K, Sourav Kanti Addya}
\IEEEauthorblockA{
Cloud \& Smart System Services Lab \\ Department of Computer Science and Engineering\\
National Institute of Technology Karnataka, Surathkal, India \\
Emails: \{hardidesai.242cs015, balajink.252is009, souravkaddya\}@nitk.edu.in
}
}


\maketitle

\begin{abstract}

Serverless inference is emerging as an efficient deployment model for Large Language Models (LLMs) due to its ability to elastically scale resources, improve GPU utilization through resource sharing, and reduce infrastructure costs. However, existing approaches primarily optimize time to first token (TTFT) while overlooking inter-token latency (ITL), a key determinant of user experience during response generation. We observe that LLM inference comprises a computation-intensive prefill stage and a bandwidth-bound decode stage, where admitting new requests can disrupt ongoing decoding and increase latency. To address this, we propose \textbf{Token Window Pacing} (\sysname), a closed-loop scheduling framework that leverages runtime feedback from ITL, queue depth, and workload characteristics to regulate request admission at the token level instead of the batch level. TWP dynamically adjusts a self-regulating token window and incorporates chunk-based prefill to minimize interference from incoming requests. Experiments on realistic serverless workloads show that TWP delivers stable TTFT, significantly reduces latency spikes, lowers P99 tail latency, minimizes ITL variation, and improves throughput by up to 45\% over ServerlessLLM and 58\% over conventional schedulers. These results demonstrate that feedback-driven token-level scheduling enables stable, high-throughput serverless LLM inference without modifying the underlying model.

\end{abstract}


\input{tex/1_intro.tex}
\input{tex/2_backgroung.tex}
\input{tex/3_prelim}
\input{tex/4_system_model}
\input{tex/5_implementation}
\input{tex/6_evaluation}

\input{tex/7_conclusion}

\bibliographystyle{IEEEtran}
\bibliography{refs}

\end{document}

%% file: tex/1_intro.tex
\section{Introduction}

Large language models (LLMs) are becoming popular for modern interactive applications such as conversational systems, programming assistants, and search engines,  where strict latency requirements make responsiveness a key component of user experience. To reduce the high cost of provisioning dedicated GPUs, many systems now rely on serverless inference, in which model instances are dynamically instantiated on demand and multiple tenants share a pooled GPU infrastructure under a pay-per-use model. It is cost-effective in that it allows service providers to optimize GPU usage via load sharing without incurring costs related to reserved capacity by end users. The downside, however, is that the practice introduces nontrivial scheduling problems that have yet to be adequately addressed in the LLM-serving literature. 

\subsection{How LLM Inference Works}

\begin{figure*}[t]
    \centering
    \includegraphics[width=0.8\linewidth]{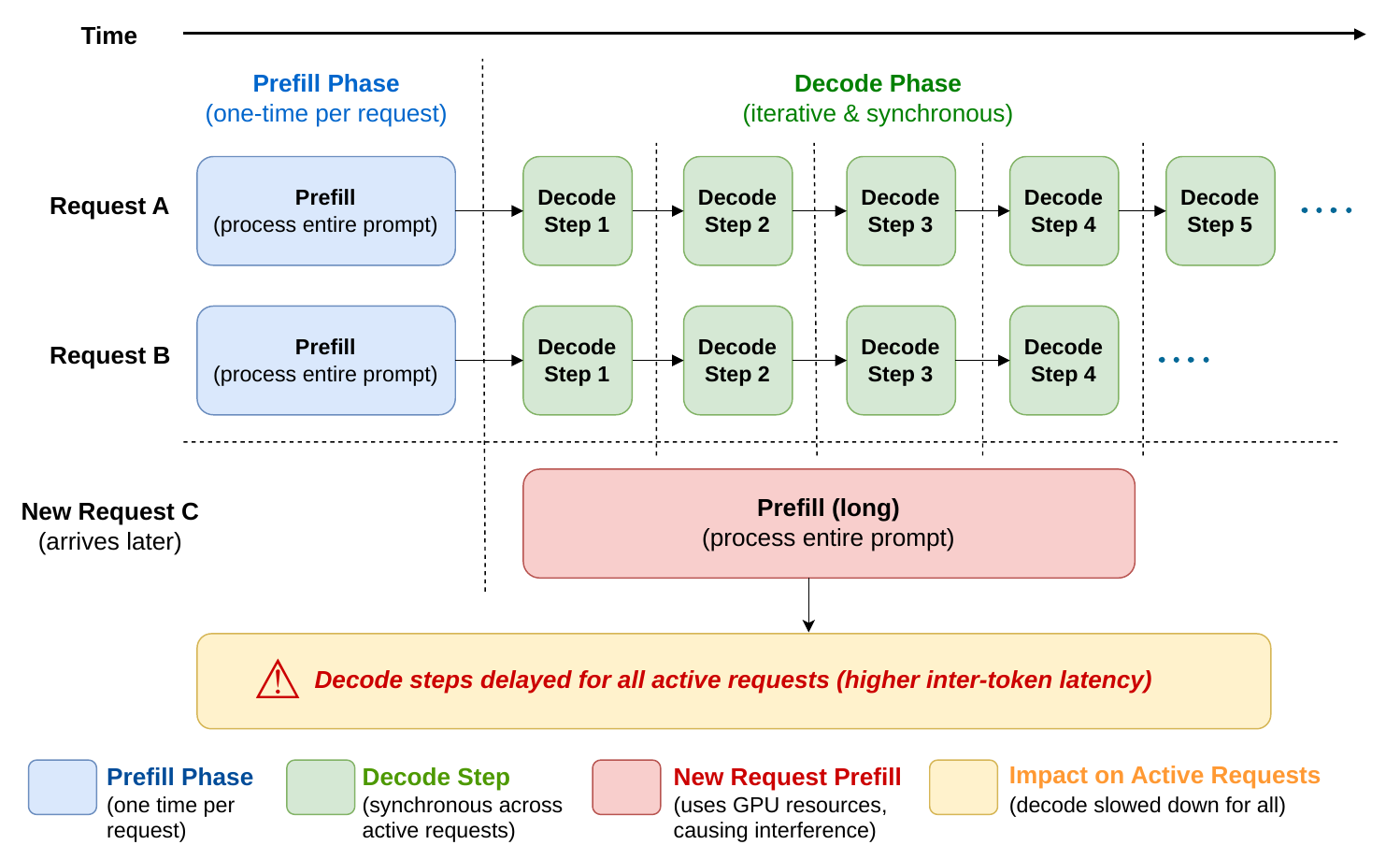}
    \caption{LLM inference pipeline showing the two-phase execution model.}
    \label{fig:llm_pipeline}
\end{figure*}

Figure~\ref{fig:llm_pipeline} illustrates the execution pipeline of LLM inference. Each request undergoes two phases: \emph{prefill} and \emph{decode}. During the \textbf{prefill} phase, the model processes the entire input prompt in a single forward pass. This step initializes the Key-Value Cache (KV-cache), which stores the attention state associated with prompt tokens and remains resident in GPU memory throughout the lifetime of the request. \emph{Prefill} executes once per request and its cost scales with prompt length.

After \emph{prefill}, the request enters the \textbf{decode} phase, where output tokens are generated iteratively, one token at a time. Decode execution is synchronous across all active requests: the system cannot advance to the next decode step until the current step completes for every request in the batch. Consequently, the latency of a decode iteration depends on the aggregate workload of the active batch, making scheduling decisions highly consequential. Admitting a new request introduces an additional prefill operation that competes with ongoing decode execution for GPU resources. The resulting interference delays token generation for all active requests, increasing inter-token latency (ITL) system-wide.

\subsection{The Admission Control Gap}

Serverless workloads are inherently bursty, with production traces showing that invocation rates can fluctuate by multiple orders of magnitude over short time intervals, and that a small fraction of functions account for a disproportionate share of total traffic \cite{sharad2020serverlessinthewild}. Such variability is particularly challenging in interactive applications, where request-rate spikes are common and can coincide across many tenants. In addition, cold-start rates exceeding 25\% in a large fraction of deployments further amplify instability in request arrival patterns~\cite{fu2024serverlessllm}. Under such circumstances, a scheduler without knowledge of the internal state of the decode engine may over-aggressively accept incoming requests during bursts, thus saturating the GPU above its efficient operating region. The result is poor efficiency in terms of serving requests, with an increase in inter-token latency (ITL), which is defined as the latency between two subsequent tokens sent to a user.

This becomes a self-reinforcing issue because higher ITL leads to slower token creation rates. This leads to increased time required for the batch to be emptied. The recovery time becomes extended, and all the users within that session experience poor stream throughput during this entire process.

A rich body of work has improved individual aspects of LLM serving. Continuous batching~\cite{yu2022orca} eliminates idle time between requests. Memory-efficient KV-cache management~\cite{kwon2023vllm} makes GPU memory use more predictable. Chunked prefill~\cite{agrawal2024sarathiserve} reduces the latency spike caused by large prompt ingestion. ServerlessLLM~\cite{fu2024serverlessllm} dramatically cuts cold-start overhead through multi-tier checkpoint loading. A Generative Scheduling framework~\cite{jin2026gensched} further advances phase-aware scheduling by jointly optimizing resource allocation for the prefill and decode stages across heterogeneous hardware. Each of these is a genuine contribution to the serving stack. Yet none of them closes the feedback loop between what the decode engine is currently experiencing and how many new requests should be admitted. The admission gate remains open-loop: decisions are made at arrival time, not in response to observed congestion.

\subsection{Our Approach}

We propose \sysname\, a scheduler that completes this feedback loop at token granularity. The core insight is that ITL, obtained at the end of every decode stage, directly and with no overhead indicates GPU decode pipeline congestion. ITL is a better indicator compared to queue depth, which can keep growing even when ITL values are low, and also more useful than KV-cache occupancy, which fails to reflect latency cost. By combining ITL and queue depth, as well as KV-cache occupancy in a proportional feedback controller, \sysname\ controls a \emph{dynamic token window} in order to ensure proactive overload prevention.

\medskip\noindent\textbf{Contributions:}
\begin{enumerate}[label=\roman*.]
   \item A \textbf{token-granularity, feedback-driven admission framework}
        for serverless LLM inference, grounded in a formal instability result showing why unconstrained admission fails under burst load~(\S\ref{sec:twp_&_arch:model}).
  \item \textbf{An ITL proof-of-concept}~(\S\ref{sec:itl}) that
        characterises ITL as a congestion signal, establishes its theoretical properties, and shows why it is the correct optimisation target for joint TTFT and decode-stability goals.
  \item The \textbf{\sysname\ scheduler}~(\S\ref{sec:impl}):
        a two-guard admission gate (token budget and KV-cache headroom), a starvation-safe composite ranking policy, and chunked prefill staging with principled parameter choices.
  \item An \textbf{evaluation}~(\S\ref{sec:eval}) against standard baselines on real workloads in a shared serverless container pool.
\end{enumerate}

%% file: tex/2_backgroung.tex
\section{Background \& Prior Works}
\label{sec:background}
This section provides the background necessary to motivate the design of \sysname. We first examine the cost and latency characteristics of LLM inference and the admission challenges that arise in shared serving environments. We then place \sysname\ in the context of prior work on serverless scheduling and LLM serving, highlighting the gap addressed by our approach.

\subsection{Cost Structure, Latency Metrics, and Admission Hazards}
\label{sec:bg:phases}
 
The introduction established that LLM inference has two phases and that the decode phase is synchronous across the batch. Here we look at the cost structure of each phase in more detail, define the two latency metrics that \sysname\ optimises, and identify two distinct admission hazards that motivate the design.
 
\textbf{Prefill cost:}
Prefill is expensive and prompt-length-dependent: every input token attends over all preceding tokens, so the compute cost scales quadratically with prompt length. Its output --- the \emph{KV-cache} --- occupies GPU memory proportional to prompt length from the moment of admission. A long prompt therefore consumes both a large burst of compute and a lasting block of memory the instant it enters the system. This is relevant for admission because both resources are shared with the ongoing decode work of already-active requests.
 
\textbf{Decode cost:}
Each decode step reads the full KV-cache of every active request and writes one new entry per request. The dominant cost is memory bandwidth: loading KV-cache tensors from GPU high-bandwidth memory (HBM). Up to a saturation batch size $B_0$, the GPU's memory bus amortises this cost across requests and step time stays flat --- batching is effectively free in this regime. Beyond $B_0$, attention computation becomes the bottleneck and step time rises with each additional request, raising ITL for every active user simultaneously. This piecewise  ost structure defines the operating region \sysname\ aims to stay within.
 
\textbf{TTFT and ITL:}
In a serverless deployment, TTFT --- the delay from request submission to the first output token --- has three components: a \emph{cold-start} penalty if no warm container is available (the model must be loaded from storage before inference can begin), followed by the prefill phase, followed by the first decode step. ITL governs the streaming experience thereafter. Formally:
\begin{equation}
  \mathrm{TTFT}_i =
    \underbrace{\mathrm{cs}_i}_{\substack{\text{cold-start}\\\text{overhead}}}
  + \underbrace{\sum_{k=t_0}^{t_0 + \lceil p_i / C \rceil} \ell_k}_{\substack{\text{chunked prefill}\\\text{in decode steps}}},
  \label{eq:ttft}
\end{equation}
where $\mathrm{cs}_i$ is the cold-start delay (zero for a warm container), $t_0$ is the admission step, and $C$ is the prefill chunk size. When prefill is staged across decode steps --- as \sysname\ does --- its cost is paid entirely in units of ITL. Minimising ITL therefore simultaneously reduces TTFT and improves streaming quality, which is why it is the right primary control target.
 
\textbf{The noisy-neighbour penalty:}
A second admission hazard arises at the moment a new request enters the system. Prefill issues a large, bursty memory operation: KV-cache entries for every input token must be written to HBM in the same step that the ongoing decode kernels are reading their own caches. This contention on the shared memory bus transiently raises ITL for all currently active requests --- an effect that scales with prompt length and is entirely invisible to a scheduler that makes decisions only at arrival time. We call this the \emph{noisy-neighbour penalty}. A long prompt admitted in one shot can temporarily double active users' ITL. Spreading ingestion across multiple steps in fixed-size chunks bounds this penalty per step to a small fraction of the full-prompt cost, at a modest TTFT cost we derive in \S\ref{sec:twp_&_arch:design:chunked}.

\subsection{The Serverless Scheduling Landscape}
\label{sec:bg:related}

We survey the existing work relevant to \sysname, organised around the distinct problems each body of work addresses, and identify the gap that remains open.

\textbf{Cold-start reduction:}
Fu et al. \cite{fu2024serverlessllm} introduce ServerlessLLM, which optimizes checkpoint loading across DRAM and NVMe memory tiers to reduce cold-start latency by a factor of 10 to 200 times. It also implements live migration that only transfers token states between servers, making serverless LLM serving practically feasible. Building on this, Cui et al. \cite{cui2025efficientfaas} present TIDAL, a solution that enhances Function as a Service (FaaS) by overlapping model loading with early inference through weight-centric tracing and adaptive state forking. However, they note that CPU-bound initialization remains unaccelerated. Yu~\etal~\cite{yu2026lambdascale} introduce {$\lambda$}Scale, whose {$\lambda$}Pipe binomial multicast pipeline constructs execution graphs across nodes during loading; the system is currently limited to single-node memory capacity. Zhu et al. \cite{zhu2025tangram} reduce repeated loading by using Tangram, which shares tensor-level parameters within a unified GPU memory pool and employs affinity-aware scheduling. This approach incurs a slight throughput overhead due to on-demand key-value cache allocation. Lou et al. \cite{lou2025hydraserve} optimize the process by overlapping model loading with container initialization through network-aware placement. Sui et al. \cite{sui2024prewarming} demonstrate that pre-warming alone is insufficient and propose opportunistic pre-loading when there is idle capacity. Additionally, Lou et al. \cite{lou2025warmserve} introduce universal GPU workers capable of pre-warming multiple models simultaneously. All of these strategies address the startup component \( \mathrm{cs}_i \) in Equation~\eqref{eq:ttft}. Once a container is warmed up, the main challenge becomes regulating the decoding batch to maintain a low inference turn-around time (ITL).

\textbf{Continuous batching and memory management:}
Orca~\cite{yu2022orca} introduces continuous batching, making the decode step the natural scheduling unit and eliminating padding waste---directly motivating \sysname's step-synchronous control. vLLM~\cite{kwon2023vllm} adds PagedAttention, managing KV-cache in fixed-size pages to eliminate fragmentation; the resulting per-step occupancy measurement underpins \sysname's second admission guard.

\textbf{Chunked prefill and phase separation:}
SARATHI~\cite{agrawal2023sarathi} and Sarathi-Serve~\cite{agrawal2024sarathiserve} split prefill into fixed-size chunks merged with decode steps, substantially reducing the noisy-neighbour penalty. These establish the chunked admission mechanism \sysname\ adopts, but neither connects observed ITL back to admission rate. Splitwise~\cite{patel2025splitwise} separates prefill and decode onto dedicated hardware pools; \sysname\ achieves phase-awareness within a single shared pool.

\textbf{SLO-aware and token-level scheduling:}
Bari~\etal~\cite{bari2025optimalscheduling} use deadline-driven criticality scheduling for TTFT and time-between-token SLOs, but require known deadline targets and target single-instance deployments. TokenScale~\cite{lai2025tokenscale} proposes Token Velocity as a predictive autoscaling metric, reinforcing that token-level signals outperform request-level metrics for capacity decisions. LAPS-SD~\cite{Li2025semiclairvoyant} combines Least-Attained-Service with SJF for speculative decoding, highlighting that output-length uncertainty is a fundamental scheduling challenge shared with \sysname. Han~\etal~\cite{han2025token} propose TALE, which dynamically adjusts chain-of-thought token budgets via zero-shot prompting or post-training internalisation; though targeting reasoning efficiency rather than serving infrastructure, it reinforces that token counts carry actionable workload-intensity signals.

\textbf{Serverless scheduling, LoRA serving, and large-scale systems:}
Thompson~\etal~\cite{thompson2026jobscheduling} derive competitive online scheduling algorithms for serverless systems with setup delays, staying within a $(1+\epsilon)$ cost factor of optimal, but assume uniform costs and make no distinction between prefill and decode phases. Ni~\etal~\cite{ni2026predictivelora} address multi-adapter cold-start with Predictive-LoRA, prefetching hot LoRA adapters via an LSTM predictor and eliminating fragmentation through page-based memory management, though accuracy degrades under dynamic patterns. Sui~\etal~\cite{sui2025serverlesslora} propose ServerlessLoRA, sharing model backbones across isolated functions via CUDA IPC with contention-aware batching and dynamic GPU offloading, at the cost of duplicated CUDA context overhead. Hu~\etal~\cite{junhao2025deepserve} present DeepServe, a production serverless platform with a microkernel serving engine and NPU-fork for fast weight transfer across nodes; specialised hardware interconnects are required and true zero-instance cold starts remain unsupported.

\textbf{Memory-constrained inference and adaptive batching:}
Alizadeh~\etal~\cite{alizadeh2024llmflash} load only sparsity-selected weights from flash into DRAM using a low-rank predictor, with windowing and row-column bundling for throughput; the work reinforces that KV-cache and weight residency must be managed jointly, a principle \sysname\ operationalises in its per-step memory guard. Ali~\etal~\cite{ali2020batch} propose BATCH, tying batch size to profiled latency targets for non-autoregressive serverless inference; the insight that batching decisions should be grounded in observed latency aligns with \sysname, though BATCH uses static thresholds and ignores KV-cache dynamics and the prefill--decode distinction.

\textbf{The open gap:}
No existing system effectively closes the feedback loop between the observed decode-phase Inference Time Latency (ITL) and the request admission gate in a shared serverless Large Language Model (LLM) pool. DynamicBatch employs a static batch-size limit without mechanisms to prevent saturation before this limit is reached. ServerlessLLM's rapid loading unintentionally leads to batch saturation by accepting requests faster than the decode phase can handle them. The root of the problem lies in the structure of the system: all existing schedulers make admission decisions at the moment a request arrives, without considering any information about the status of the current batch from the decode loop. Our system, \sysname, addresses this issue by making admission decisions during the decode step, using the ITL measured in the previous step to inform these choices.

%% file: tex/3_prelim.tex
\section{ITL as a Congestion Signal}
\label{sec:itl}

We establish the formal and empirical case for ITL as the primary feedback signal in \sysname. All characterisation in this section is derived from measurements on our experimental platform.

\subsection{Definition and User-Visible Meaning}

\begin{definition}[Inter-Token Latency]
Let $t_k$ be the wall-clock time at which decode step $k$ completes. The \emph{inter-token latency} at step $k$ is $\ell_k = t_k - t_{k-1}$. \end{definition}

Because the decode loop is synchronous, every active request experiences the same $\ell_k$ at each step. From a user's perspective, $\ell_k$ is the gap between successive tokens in the response stream. A stable, low $\ell$ produces smooth streaming; a high or volatile $\ell$ produces visible stuttering regardless of how fast the total response eventually arrives.

\subsection{The ITL--Batch-Size Relationship}
\label{sec:itl:model}

The duration of a decode step is governed by either KV-cache transfers from GPU high-bandwidth memory (HBM) or attention computation. For small batches, memory transfers dominate and the GPU can load KV-caches for multiple requests without increasing total transfer time. Consequently, step duration remains constant at a minimum value $\itlmin$ until the batch size reaches a saturation point $B_0$.

\begin{figure}[h]
\centering
\includegraphics[height=5cm, width=\linewidth]{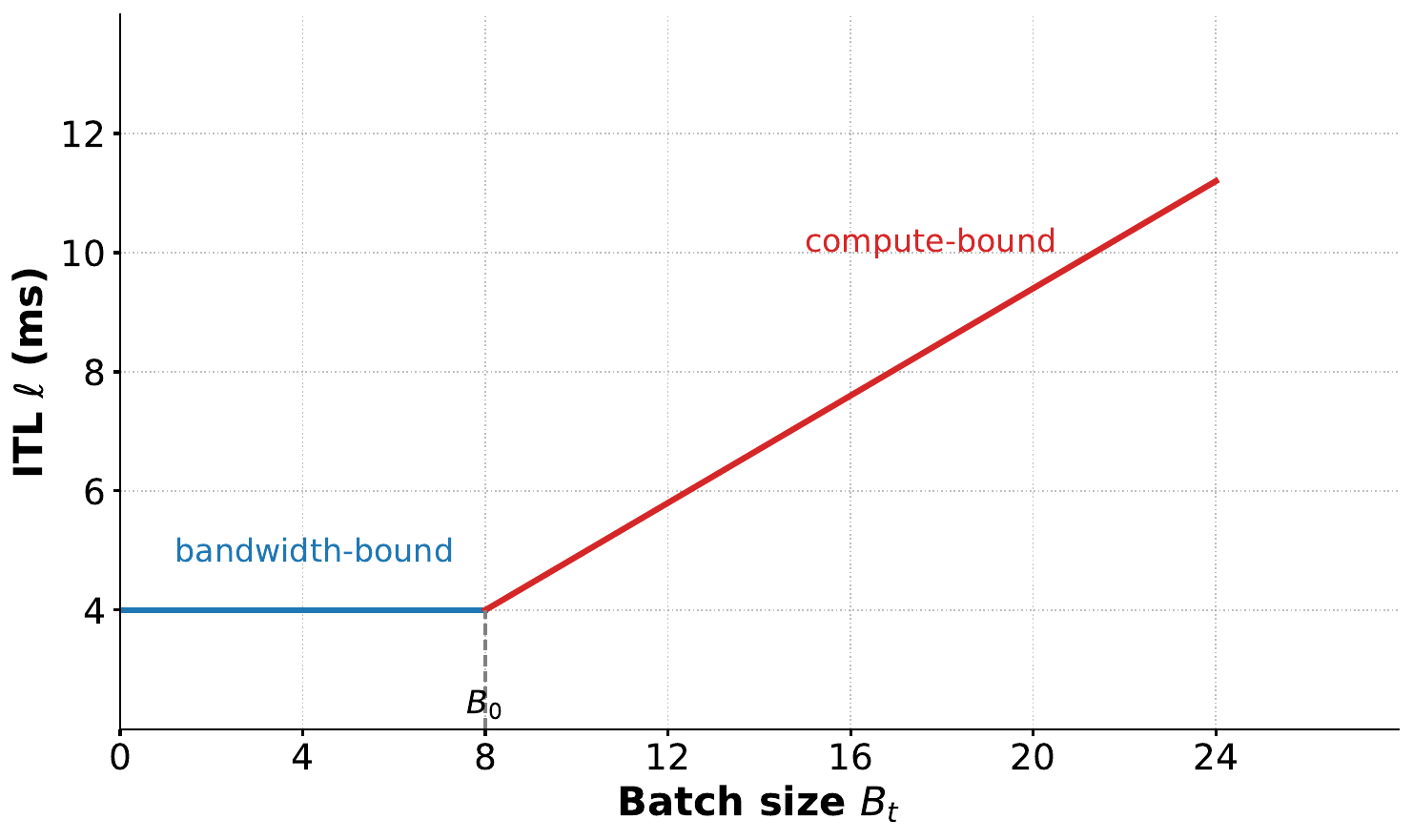}
\caption{TTFT comparison across different scheduling policies.}
\label{fig:hw_itl}
\end{figure}

Beyond $B_0$, attention computation becomes the bottleneck. Each additional request increases the computation required to process its KV-cache, causing step duration to grow approximately linearly with batch size. The resulting ITL model is

\begin{equation}
  \ell(B_t) =
  \begin{cases}
    \itlmin, & B_t \leq B_0, \\
    \itlmin + \alpha(B_t - B_0), & B_t > B_0.
  \end{cases}
  \label{eq:itl_model}
\end{equation}

where $B_0$ denotes the transition point between bandwidth-bound and compute-bound execution, and $\alpha$ captures the marginal increase in decode-step latency per additional request. These parameters are obtained through a one-time offline profiling sweep that measures step duration across different batch sizes. Figure~\ref{fig:hw_itl} illustrates the characteristic behaviour and the inflection at $B_0$.

\subsection{Instability Without Admission Control}

At steady state, tokens arrive at rate $\lambda\bar{L}$ and are processed at rate $B^*/\ell(B^*)$. For $B^*>B_0$, the service rate satisfies $B^*/\ell(B^*) \rightarrow 1/\alpha$ as $B^*\rightarrow\infty$. Thus, the maximum achievable token processing rate is bounded by $1/\alpha$.

When $\lambda\bar{L} > 1/\alpha$, the token arrival rate exceeds the maximum service rate. The backlog therefore grows continuously, implying that no finite steady-state batch size exists. In the absence of an explicit bound on token load, $B^*$ diverges without limit.

Even when $\lambda\bar{L} < 1/\alpha$, transient bursts can temporarily raise the effective arrival rate above $1/\alpha$. During such periods, the system enters the compute-bound regime, causing the active batch size to grow. Recovery occurs only after the excess queued requests are processed, which may require many decoding steps. As a result, all active users experience elevated ITL throughout the recovery period. Admission control through a token window prevents unbounded batch growth and limits the duration and severity of these overload episodes.

\subsection{Why ITL Dominates Alternative Signals}

\textbf{Queue depth} reflects demand but not congestion. The queue can grow while $B_t < B_0$ and ITL remains at $\itlmin$. Throttling on queue depth alone would delay admission during periods where the system has available headroom.

\textbf{KV-cache occupancy} reflects memory pressure but not compute pressure. A batch of many short-output requests may have high occupancy while still residing comfortably in the bandwidth-bound regime. Memory occupancy is a necessary guard against hard out-of-memory failures but is not sufficient as the sole control signal.

\textbf{ITL} responds within one decode step to any change in $B_t$.
It requires no additional measurement infrastructure: the step wall-clock time is already tracked by the serving engine as a standard bookkeeping value. It directly encodes the cost being imposed on every active user by the current batch composition. And from Equation~\eqref{eq:ttft}, it is the quantity that both TTFT and streaming latency are built from --- making it the single most actionable signal for an admission controller that cares about both metrics.

%% file: tex/4_system_model.tex
\section{\sysname\ and System Architecture}
\label{sec:twp_&_arch}
This section presents the \sysname\ framework and the overall architecture. We first formalise the system model and the token-based control mechanism used to regulate admission pressure during inference. We then describe the architectural components of the system and how they interact to adapt scheduling decisions based on runtime feedback.

\subsection{System Model}
\label{sec:twp_&_arch:model}
\sysname\ models the serving system as a discrete-time process advancing one decode step at a time. Each request $i$ carries a prompt length $p_i$ and a target output length $L_i$. Once admitted, it joins the \emph{active set} $A_t$ and generates one token per step until complete.

\begin{definition}[Token Load and Batch Size]
The \emph{token load} $K_t = \sum_{i \in A_t} r_i(t)$ is the total remaining output tokens across all active requests at step $t$. The \emph{batch size} $B_t = |A_t|$ is the count of active requests.
\end{definition}

Token load is the quantity \sysname\ controls. Unlike batch size, which counts requests, token load captures how much work remains: a batch of requests near completion imposes little future pressure, while the same batch size with requests just admitted imposes far more. Bounding $K_t$ therefore gives a more precise handle on where the system sits relative to saturation than bounding $B_t$ directly.

\sysname\ maintains a dynamic token budget $H_t$, referred to as the \emph{token window}, as a soft upper bound on token load such that $K_t < H_t$. Since $B_t \approx K_t / \bar{r}_t$, keeping $K_t$ below $H_t$ indirectly keeps $B_t$ below the saturation threshold $B_0$ and thereby keeps $\ell(B_t)$ close to $\itlmin$.

The window is not fixed. It adapts each step using the normalised lowdown $\sigma_t = \hat\ell_t / \itlmin$, which measures how far the current ITL has risen above the efficient baseline. The target window is:
\begin{equation}
  H_t^{*} = H_0 \cdot
    \frac{1 + \min(1,\,q_t / Q)}
         {\max(0.5,\,\sigma_t)},
  \label{eq:target}
\end{equation}
where $q_t$ is current queue depth and $H_0$ is the nominal window set at system startup. When ITL is low and the queue is growing, the window expands to absorb more work. When ITL rises above baseline, the window contracts to reduce admission pressure. The window is updated smoothly each step:
\begin{equation}
  H_{t+1} = H_t + \gamma(H_t^{*} - H_t),
  \label{eq:update}
\end{equation}
with gain $\gamma$ controlling how quickly the window tracks its target.

\subsection{System Design}
\label{sec:twp_&_arch:design}
This subsection presents the architectural design of \sysname\ and the interaction between its core components. We describe the overall system organisation, the feedback control mechanism and the per-step scheduling process that governs request execution during inference.

\subsubsection{Architecture}
\label{sec:twp_&_arch:design:arch}

\begin{figure*}[t]
    \centering
    \includegraphics[width=\textwidth]{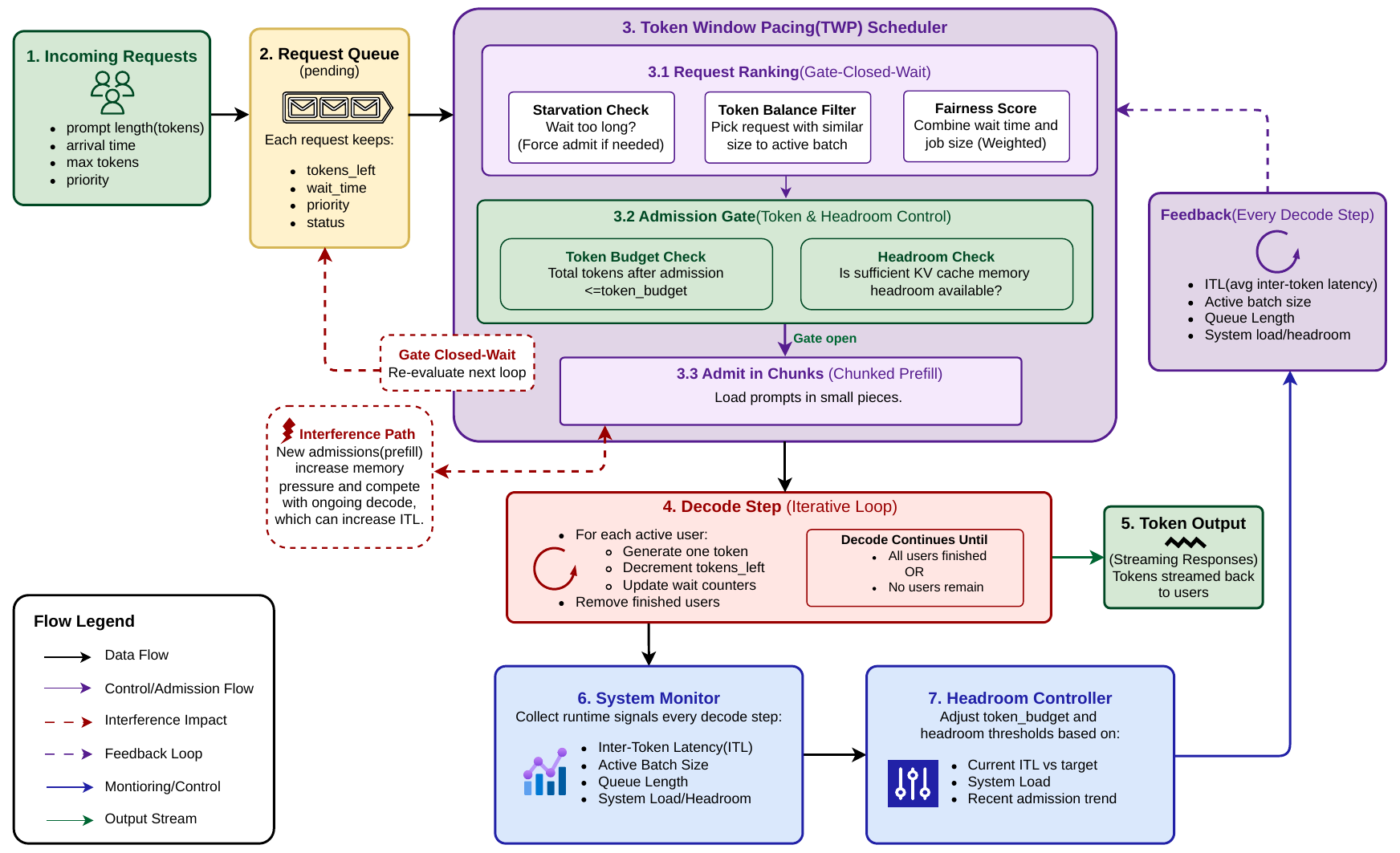}
    \caption{System-level design of the Token Window Pacing (TWP) scheduler. The scheduler regulates token-level admission using feedback from decode execution.}
    \label{fig:arch}
\end{figure*}

Figure~\ref{fig:arch} shows the \sysname\ architecture. Incoming requests queue until the scheduler admits them; admitted requests move to the decode engine, which generates one token per active request per step and streams output to clients. After each step, a monitor reads the step latency and memory state, and a headroom controller uses these to update the token window before the next step begins. This feedback path is the core contribution: the system continuously observes the cost of the current batch and adjusts admission pressure accordingly.

A key property is that the entire feedback path executes within the decode-step boundary: the monitor reads after step $t$ completes, the controller updates $H_{t+1}$, and the scheduler uses $H_{t+1}$ for the admissions that precede step $t{+}1$. There is no asynchronous thread or polling delay. Every admission decision is informed by the state one step ago, which is the most recent state available.

All schedulers in the evaluation share the same serverless container pool. Containers lose warmth over idle time, and a cold container incurs a startup delay before serving its first request. The token window operates only within the decode phase of running containers and does not interact with container provisioning.

\subsubsection{Feedback Control}
\label{sec:twp_&_arch:design:monitor}

The monitor maintains a smoothed ITL estimate $\hat\ell_t$ by applying an exponential filter to the raw step latency, attenuating transient spikes from events unrelated to batch load while still responding to genuine load changes within a few steps. The headroom controller uses $\hat\ell_t$ to compute $\sigma_t$ and update the token window via Equations~\eqref{eq:target}--\eqref{eq:update}.

In addition to ITL, the controller tracks the rate of KV-cache block consumption. When memory is being depleted rapidly, the KV-cache admission threshold $\delta_\mathrm{kv}$ is tightened to slow intake before the engine runs out of pages mid-step. When memory is stable, it relaxes. This mechanism operates on the memory signal independently of the ITL signal; both can act simultaneously.

The token window is clamped between a minimum and maximum value. The minimum ensures the system always keeps enough requests active to use available GPU parallelism; the maximum limits how far the window can expand during a quiet period, preserving headroom to absorb a subsequent burst without crossing $B_0$.

\subsubsection{Per-Step Scheduling}
\label{sec:twp_&_arch:design:scheduler}

At each decode step, the scheduler ranks all pending requests, evaluates
the admission gate, and stages admitted requests via chunked prefill.

\begin{enumerate}[label=\roman*.]
  \item \textbf{Request Ranking:} Pending requests are sorted by a composite score that balances three objectives. First, a \emph{starvation priority} proportional to how long a request has waited: once a request exceeds a maximum wait threshold $T_\mathrm{max}$, it is force-admitted unconditionally, bypassing the gate. This gives a hard worst-case latency bound regardless of window pressure. Second, a \emph{token balance} score that favours requests whose expected output length is close to the mean remaining length of the active batch, expressed as:
\begin{equation}
  T(r) = \exp\!\left(-\frac{|\hat{L}(r) - \bar{r}_t|}{\bar{r}_t + 1}\right).
\end{equation}
Admitting requests that match the current batch composition reduces the fraction of steps where a few long-running requests hold up an otherwise finished batch. Third, a \emph{fairness score} that normalises wait time by job size, preventing a stream of short requests from indefinitely displacing larger ones. The three scores are combined with weights $w_1 > w_2 > w_3$, prioritising the starvation guarantee first, batch efficiency second, and fairness third.

  \item \textbf{Admission Gate:} Candidates are considered in ranked order. A request is admitted only if two conditions hold simultaneously: the resulting token load stays within the window ($K_t + r_i \leq H_t$), and available KV-cache memory exceeds the threshold $\delta_\mathrm{kv}$. If either condition fails, the gate closes and remaining candidates wait for the next step. The gate does not skip to smaller requests that might fit within residual headroom, because doing so would bypass the starvation ordering. Force-admitted requests are exempt from both checks.
\item \label{sec:twp_&_arch:design:chunked}\textbf{Chunked Prefill:} Admitted requests do not immediately submit their full prompt for prefill. Instead, the prompt is broken into fixed-size chunks processed one per decode step. This bounds the noisy-neighbour penalty per step to a fixed fraction of $\itlmin$ regardless of prompt length. The chunk size $C$ is chosen at the point where further reduction yields negligible additional ITL improvement while TTFT continues to increase --- identified during offline characterisation as described in \S\ref{sec:impl}.
\item \textbf{Algorithm:} Algorithm~\ref{alg:twp} shows the complete per-step execution. The five steps run in strict sequence so that each admission decision uses the most recently measured system state.

\begin{algorithm}[t]
\caption{Token Window Pacing (TWP) Scheduler}
\label{alg:twp}
\begin{algorithmic}[1]
\State Initialize token window $H_0$
\State Initialize active set $A \gets \emptyset$

\For{each decode step $t$}

    \State \textbf{// Step 1: Monitor system state}
    \State Measure inter-token latency $\ell_t$
    \State Observe queue length $q_t$ and batch size $B_t$

    \State \textbf{// Step 2: Update token window (feedback control)}
    \State Compute slowdown $\sigma_t = \ell_t / \ell_{\min}$
    \State Compute target window $H^*$ based on $\sigma_t$ and $q_t$
    \State $H_{t} \gets H_{t-1} + \gamma (H^* - H_{t-1})$

    \State \textbf{// Step 3: Request selection}
    \State Rank queued requests using:
    \State \quad (i) starvation priority (waiting time)
    \State \quad (ii) token balance (similar size to active set)
    \State \quad (iii) fairness score

    \State \textbf{// Step 4: Admission control}
    \While{$K_t < H_t$ and queue not empty}
        \State Select next request $r$
        \If{adding $r$ does not exceed token window}
            \State Admit $r$ (possibly in chunks)
            \State Add $r$ to active set $A$
        \Else
            \State \textbf{break}
        \EndIf
    \EndWhile

    \State \textbf{// Step 5: Decode execution}
    \For{each request $i \in A$}
        \State Generate one token
        \State $r_i \gets r_i - 1$
    \EndFor
    \State Remove completed requests from $A$
    \State Update total tokens $K_t = \sum_{i \in A} r_i$

\EndFor
\end{algorithmic}
\end{algorithm}
\end{enumerate}

%% file: tex/5_implementation.tex
\section{Implementation}\label{sec:impl}

\sysname\ is implemented as a scheduling layer between the request queue and an existing LLM serving backend. At each decode step, it observes the step latency, active batch size, and KV-cache occupancy, updates the token window, and determines which queued requests to admit before the next step. The backend remains unchanged and continues to handle model execution and KV-cache management. As a result, \sysname\ can be integrated into any serving framework that exposes a step-level decode interface, without modifications to model weights, kernels, or GPU execution primitives.

Table~\ref{tab:notation} summarises the notation used throughout the design.

\begin{table}[h]
\centering
\caption{Notation used in \sysname.}
\label{tab:notation}
\small
\begin{tabular}{ll}
\toprule
\textbf{Symbol} & \textbf{Meaning} \\
\midrule
$H_t$              & Token window at step $t$ \\
$H_0$              & Initial token window \\
$K_t$              & Total tokens remaining across active requests \\
$B_t$              & Active batch size \\
$\itlmin$          & Baseline decode step latency (bandwidth-bound) \\
$B_0$              & Saturation batch size \\
$\sigma_t$         & Normalised ITL slowdown $= \hat\ell_t / \itlmin$ \\
$\gamma$           & Token window control gain \\
$C$                & Prefill chunk size (tokens) \\
$q_t$              & Queue depth at step $t$ \\
$T_{\max}$         & Starvation force-admit threshold (steps) \\
$\hat\ell_t$       & Smoothed ITL estimate \\
\bottomrule
\end{tabular}
\end{table}

\sysname\ requires three deployment-time parameters. The initial token window is set to $H_0=B_0\bar{L}$, where $B_0$ is the measured saturation batch size and $\bar{L}$ is the expected mean output length. The control gain $\gamma$ determines the responsiveness of the feedback controller, while the prefill chunk size $C$ controls the trade-off between TTFT and decode interference. In our implementation, we use $H_0=400$ tokens, $\gamma=0.15$, and $C=64$ tokens. These values are obtained through offline characterization and remain fixed throughout all experiments.

\sysname\ controls request admission, batching, and execution ordering, but does not modify the underlying model architecture or execution stack. This design enables deployment on existing LLM serving systems with minimal integration effort.

%% file: tex/6_evaluation.tex
\section{Evaluation}\label{sec:eval}

We evaluate \sysname\ against conventional scheduling policies and \textit{ServerlessLLM}~\cite{fu2024serverlessllm} to assess its ability to reduce latency while maintaining throughput under bursty workloads.

\subsection{Experimental Setup}

\textbf{Implementation and Environment:}
\sysname\ is implemented in Python as a scheduling layer integrated into a custom LLM serving pipeline supporting request-level and token-level scheduling. Experiments are conducted on NVIDIA Tesla V100 GPUs (32,GB) with CUDA 11.8. Unless otherwise stated, all results use a single GPU serving the Meta-Llama-3.2-3B model~\cite{llama3.2} in FP16 precision.

\textbf{Workload:}
Requests are generated from the GSM8K benchmark~\cite{cobbe2021gsm8k}. Arrivals follow a Poisson process with rate $\lambda=18$ req/s over a 60-second interval, producing up to 250 requests per run. Prompt lengths are heterogeneous: 60\% short (30--80 tokens), 30\% medium (120--250 tokens), and 10\% long (350--512 tokens). The saturation batch size is $B_0=8$.

\textbf{Baselines:}
We compare \sysname\ against representative scheduling policies spanning classical queueing, batching, and serverless execution strategies. ServerlessLLM is reimplemented within our runtime to ensure a fair comparison under identical resource and execution conditions. Table~\ref{tab:baselines} summarizes the evaluated schedulers.

\begin{table}[h] 
\centering 
\caption{Baseline schedulers and their admission strategies.} \label{tab:baselines} 
\small \begin{tabular}{lp{5.0cm}} 
\toprule 
\textbf{Scheduler} & \textbf{Policy} \\ 
\midrule 
FCFS~\cite{zhao1989fcfs} & Admit in arrival order; no concurrency bound. \\ 
Round-Robin~\cite{hahne1991roundrobin} & Route requests to containers in rotation. \\ 
Priority~\cite{Lehoczky1991priority} & Admit in shortest-prompt-first order. \\ 
DynamicBatch~\cite{liu2007dynamicbatch} & Cap batch size when ITL exceeds a static target threshold; no token-level feedback. \\ 
ServerlessLLM~\cite{fu2024serverlessllm} & Implemented within our runtime using FCFS-style admission with locality-aware startup scheduling, fast checkpoint loading, and live migration support. \\ \\ 
\bottomrule 
\end{tabular} 
\end{table} 

\textbf{Metrics:}
We report Time-to-First-Token (TTFT), end-to-end latency (P50/P95/P99), throughput (tokens/s), and inter-token latency (ITL). To quantify generation stability, we additionally report the ITL coefficient of variation (CV).

\subsection{Results}

This section presents the experimental results obtained from the evaluation setup described earlier. Using the defined baselines and evaluation metrics, we compare \sysname\ against existing scheduling policies across various metrics and further analyse its behaviour under diverse workload patterns.

\subsubsection{Time-To-First-Token (TTFT)}

\begin{figure}[H]
    \centering
    \includegraphics[width=\linewidth]{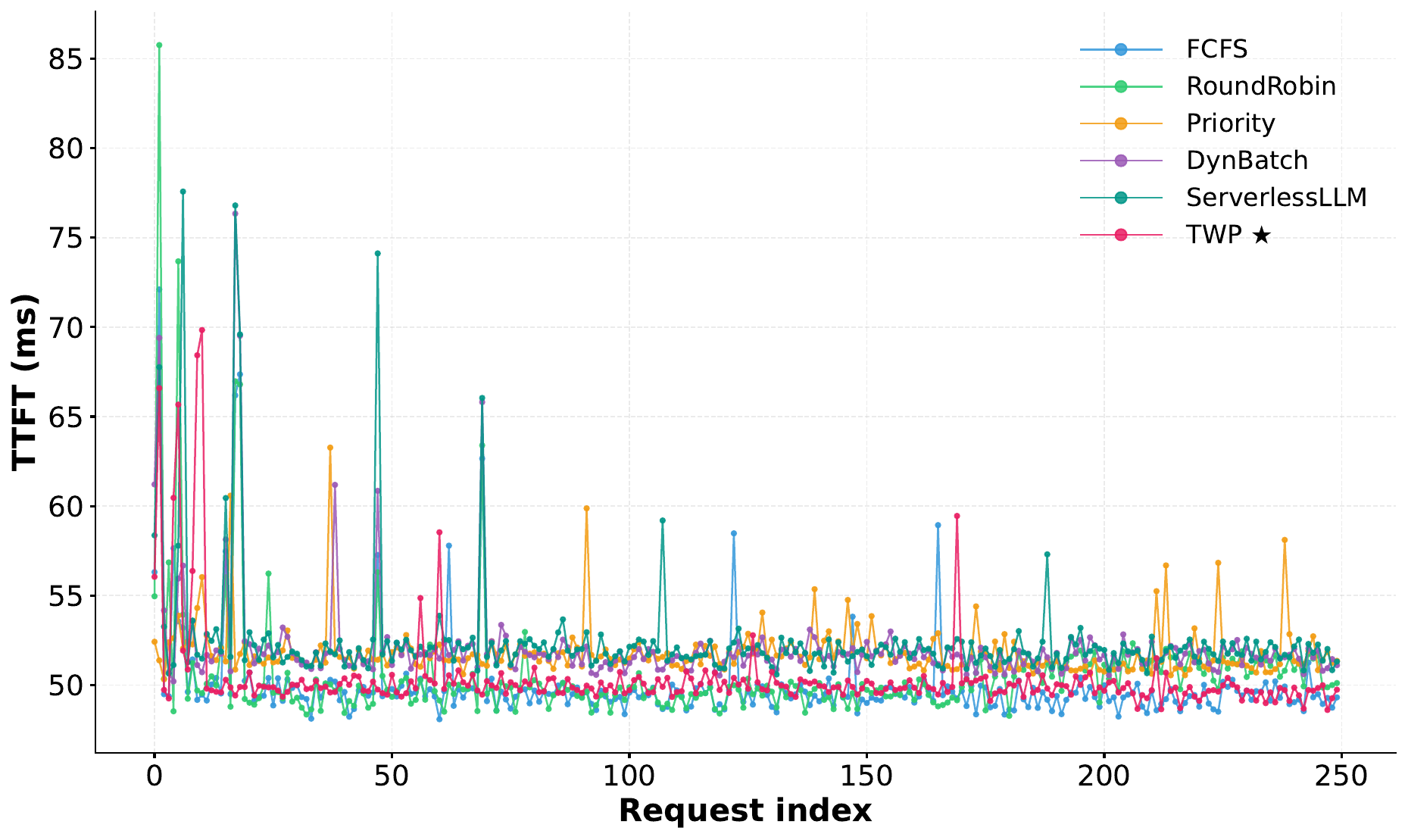}
    \caption{TTFT comparison across different scheduling policies.}
    \label{fig:ttft}
\end{figure}

Figure~\ref{fig:ttft} shows TTFT for individual requests. While Round Robin and ServerlessLLM experience latency spikes of up to 85\,ms and 77\,ms, respectively, \sysname\ maintains TTFT near 50\,ms throughout the workload. This improvement results from feedback-driven admission control, which limits queue buildup during transient bursts.

\subsubsection{TTFT Tail Latency}

\begin{figure}[H]
    \centering
    \includegraphics[width=\linewidth]{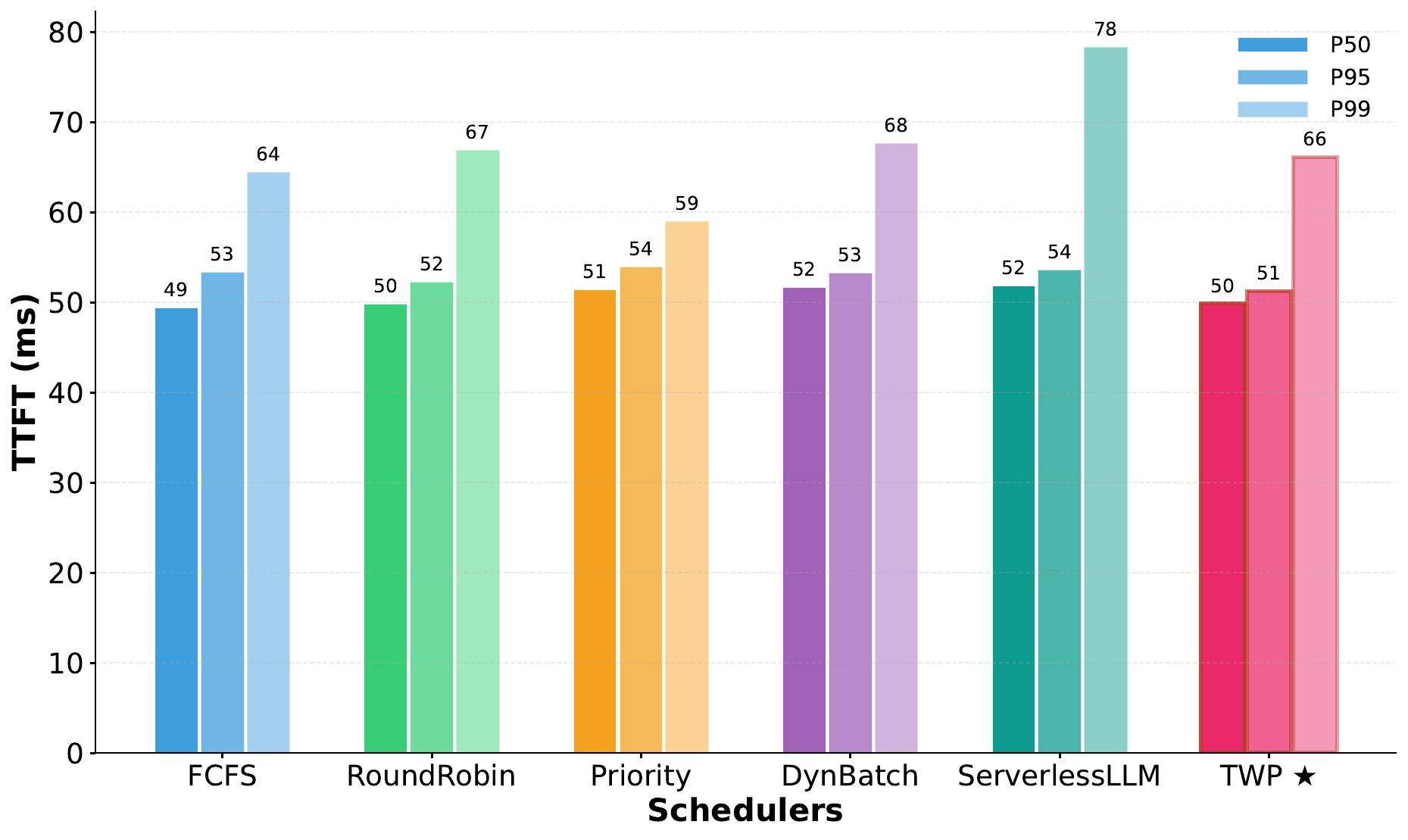}
    \caption{Comparison of TTFT percentiles (P50, P95, and P99) across different scheduling policies.}
    \label{fig:p99}
\end{figure}

Figure~\ref{fig:p99} reports TTFT percentiles. Although all schedulers achieve comparable median latency (P50 $\approx$ 50\,ms), larger differences emerge at the tail. \sysname\ achieves P95 and P99 latencies of 51\,ms and 66\,ms, compared to 78\,ms P99 for ServerlessLLM and 68\,ms for DynamicBatch, indicating improved robustness under contention.

\subsubsection{Throughput}

\begin{figure}[H]
    \centering
    \includegraphics[width=\linewidth]{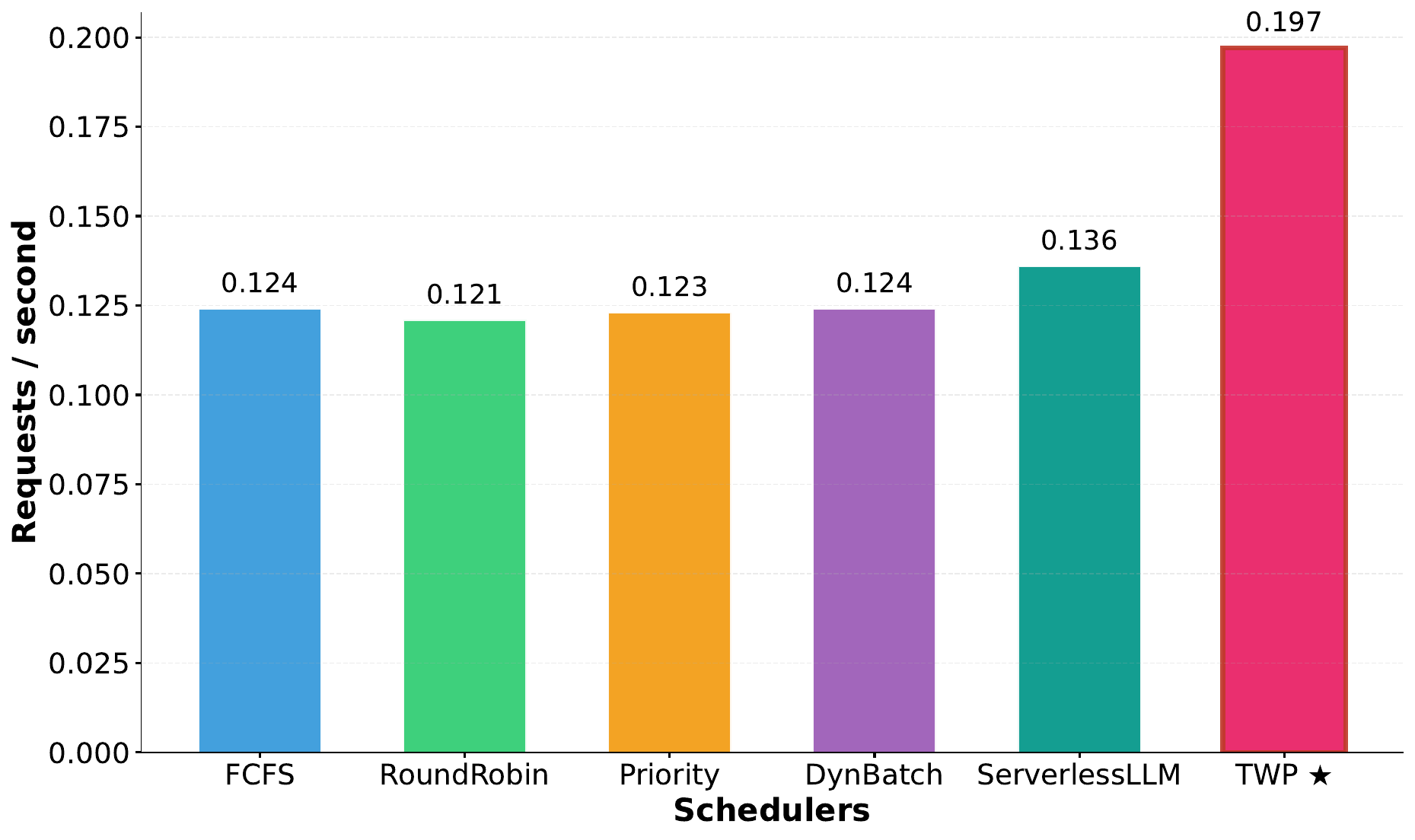}
    \caption{Throughput comparison across different scheduling policies.}
    \label{fig:throughput}
\end{figure}

Figure~\ref{fig:throughput} compares throughput across schedulers. Traditional policies achieve 0.121--0.124 requests/second, while ServerlessLLM reaches 0.136 requests/s. \sysname\ achieves the highest throughput at 0.197 requests/s, representing gains of approximately 45\% over ServerlessLLM and more than 58\% over conventional schedulers. The improvement stems from preventing excessive batch growth while maintaining high GPU utilization.

\subsubsection{Inter-Token Latency (ITL)}

\begin{figure}[H]
    \centering
    \includegraphics[width=\linewidth]{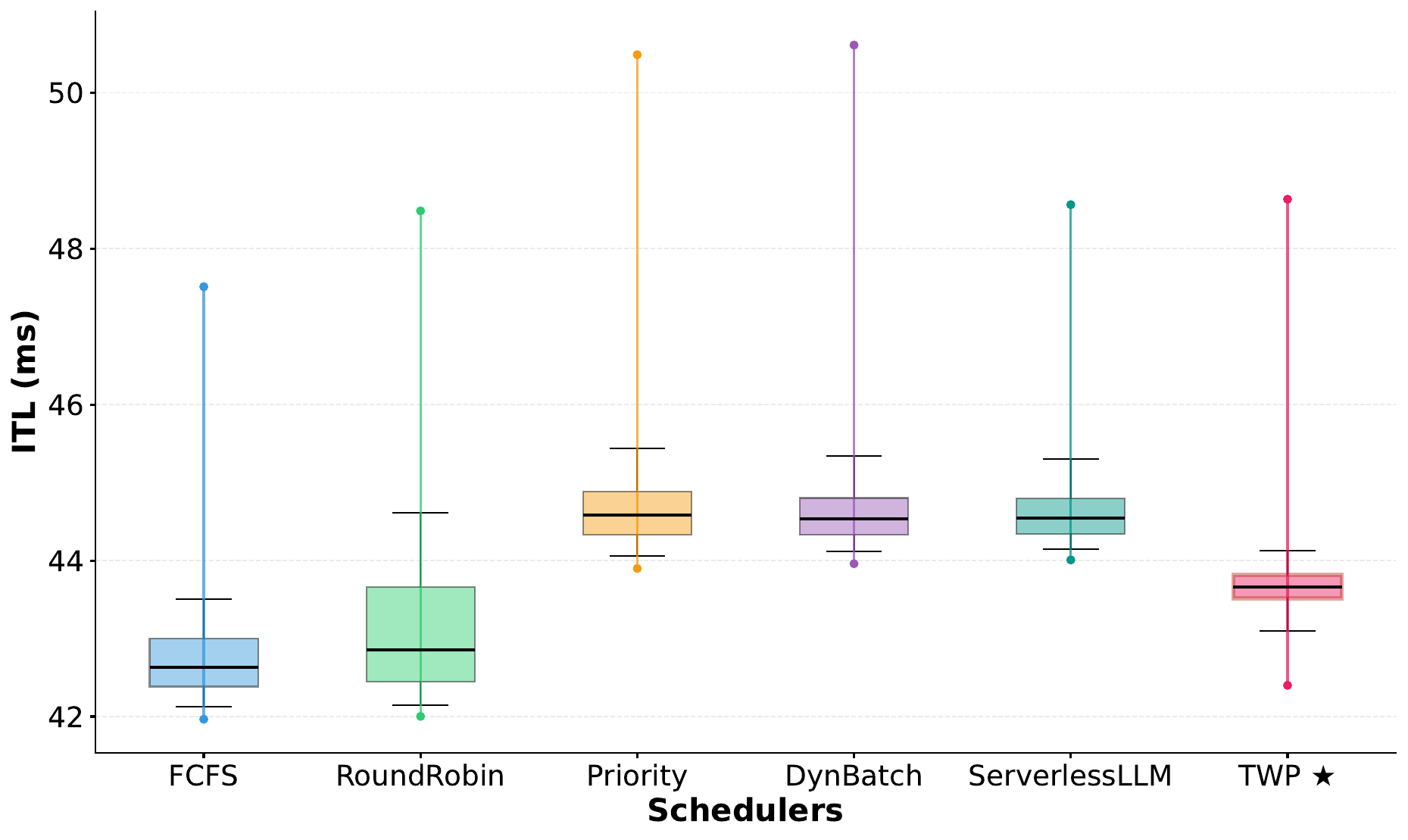}
    \caption{ITL distribution across different scheduling policies.}
    \label{fig:itl_box}
\end{figure}

Figure~\ref{fig:itl_box} shows the ITL distribution. Baseline schedulers exhibit larger variance and more frequent high-latency outliers, with tail values reaching 61--62\,ms. In contrast, \sysname\ maintains a tighter distribution centered around 43.5--43.8\,ms, indicating more stable token generation under concurrent workloads.

\subsubsection{ITL Stability}

\begin{figure}[H]
    \centering
    \includegraphics[width=\linewidth]{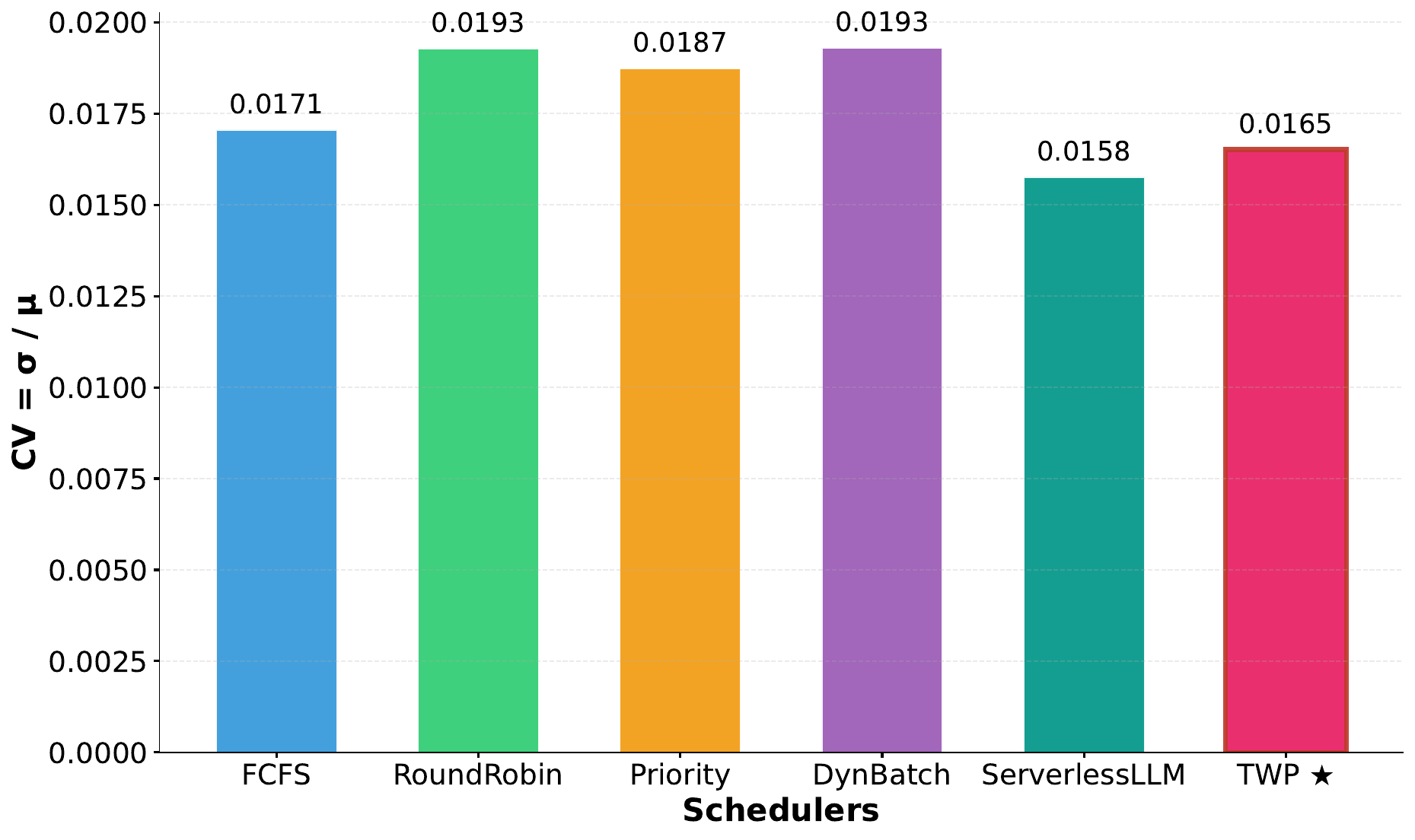}
    \caption{Coefficient of variation (CV) of inter-token latency across scheduling policies.}
    \label{fig:cv}
\end{figure}

Figure~\ref{fig:cv} reports the coefficient of variation of ITL. \sysname\ achieves the lowest variability (CV = 0.0144), compared to 0.0203 for Round Robin and 0.0159 for FCFS, demonstrating more consistent decoding performance.

\subsubsection{Evaluation Under Diverse Workload Patterns}

To evaluate robustness under varying load conditions, we consider four workload scenarios: \textit{low\_load}, \textit{medium\_load}, \textit{high\_load}, and \textit{burst\_load}, summarized in Table~\ref{tab:allloads}. Each scheduler is evaluated over 10 independent runs per workload.

Request traces use heterogeneous prompt and output lengths. Prompt lengths are sampled from {50, 200, 800} tokens with probabilities {0.5, 0.3, 0.2}, while output lengths are sampled from {20, 50, 200} tokens with the same probabilities. This setup captures both short interactive requests and long-context generation workloads.

\begin{table}[h]
\centering
\caption{Performance across workloads}
\label{tab:allloads}
\footnotesize
\setlength{\tabcolsep}{2.8pt}
\begin{tabular}{lccccc}
\toprule
\textbf{Scheduler} & \textbf{Throughput} & \textbf{P99 Latency} & \textbf{TTFT} & \textbf{ITL} & \textbf{CV ITL} \\
& (req/s) & (ms) & (ms) & (ms) & \\
\midrule

\multicolumn{6}{c}{\textit{Low Load}} \\
\midrule
FCFS          & 363.09  & 2130.93  & 52.99 & 44.29 & 0.0141 \\
DynamicBatch  & 363.09  & 2130.93  & 52.17 & 43.26 & 0.0145 \\
Priority      & 433.87  & 2134.36  & 52.20 & 43.47 & 0.0142 \\
RoundRobin    & 334.48  & 2147.98  & 52.17 & 43.30 & 0.0147 \\
ServerlessLLM & 451.32  & 2014.46  & 51.99 & 43.19 & 0.0142 \\
\textbf{\sysname}  & \textbf{2568.31} & \textbf{1152.40} & \textbf{52.02} & \textbf{43.11} & \textbf{0.0141} \\
\midrule

\multicolumn{6}{c}{\textit{Medium Load}} \\
\midrule
FCFS          & 267.13  & 4571.94  & 52.47 & 44.05 & 0.0140 \\
DynamicBatch  & 267.13  & 4269.44  & 51.47 & 43.02 & 0.0129 \\
Priority      & 467.47  & 4527.29  & 52.35 & 44.01 & 0.0144 \\
RoundRobin    & 254.80  & 4618.15  & 52.28 & 43.85 & 0.0138 \\
ServerlessLLM & 313.88  & 4263.09  & 52.16 & 43.69 & 0.0137 \\
\textbf{\sysname}  & \textbf{2674.39} & \textbf{2594.39} & \textbf{52.09} & \textbf{43.09} & \textbf{0.0136} \\
\midrule

\multicolumn{6}{c}{\textit{High Load}} \\
\midrule
FCFS          & 254.62  & 8915.61  & 51.94 & 43.48 & 0.0141 \\
DynamicBatch  & 254.62  & 8493.63  & 51.89 & 43.41 & 0.0131 \\
Priority      & 448.65  & 8833.61  & 53.37 & 45.09 & 0.0200 \\
RoundRobin    & 242.44  & 8571.43  & 52.08 & 43.56 & 0.0136 \\
ServerlessLLM & 299.29  & 8005.79  & 53.42 & 45.14 & 0.0173 \\
\textbf{\sysname}  & \textbf{2709.57} & \textbf{5103.69} & \textbf{52.86} & \textbf{44.10} & \textbf{0.0146} \\
\midrule

\multicolumn{6}{c}{\textit{Burst Load}} \\
\midrule
FCFS          & 226.30  & 17869.95 & 53.25 & 44.95 & 0.0169 \\
DynamicBatch  & 226.30  & 16894.07 & 53.13 & 44.83 & 0.0172 \\
Priority      & 452.60  & 17488.25 & 53.34 & 45.22 & 0.0177 \\
RoundRobin    & 219.97  & 18052.02 & 53.38 & 45.03 & 0.0172 \\
ServerlessLLM & 258.57  & 16915.02 & 53.35 & 45.04 & 0.0179 \\
\textbf{\sysname}  & \textbf{2532.53} & \textbf{10272.56} & \textbf{52.49} & \textbf{44.68} & \textbf{0.0169} \\
\bottomrule
\end{tabular}
\end{table}

Across all workloads, \sysname\ consistently achieves lower latency and higher throughput than the baselines. The gains become more pronounced under high-load and burst-load conditions, where uncontrolled admission leads to queue buildup and increased decode interference. By dynamically regulating token admission, \sysname\ maintains stable TTFT and ITL even under sustained bursts.

%% file: tex/7_conclusion.tex
\section{Conclusion}
\label{sec:conclusion}

This work presented (\sysname), a feedback-driven scheduling layer for serverless LLM inference that regulates request admission at decoding-step granularity using only runtime signals available in existing serving systems. \sysname\ operates without modifying model architectures, attention kernels, KV-cache management, or GPU execution primitives, making it broadly deployable.

Experiments on Poisson-driven workloads in a shared serverless environment show that \sysname\ consistently outperforms existing scheduling policies and ServerlessLLM. It achieves up to 58\% higher throughput, reduces tail TTFT by 15--20\%, and provides the lowest inter-token latency variability among all evaluated methods. These results demonstrate that step-level, feedback-driven token admission is an effective mechanism for stabilizing LLM inference under dynamic concurrency.